\documentclass[11pt]{article}

\usepackage[final]{acl}

\usepackage{times}
\usepackage{latexsym}

\usepackage[T1]{fontenc}

\usepackage[utf8]{inputenc}

\usepackage{microtype}

\usepackage{inconsolata}

\usepackage{graphicx}

\usepackage{hyperref}
\usepackage{amsmath,amssymb,amsfonts}
\usepackage{algorithmic}
\usepackage{graphicx}
\usepackage{textcomp}
\usepackage{enumitem} 
\usepackage{array}
\usepackage[most]{tcolorbox}
\usepackage{stfloats}
\usepackage{booktabs}
\usepackage{multirow}
\usepackage{placeins}
\usepackage{tabularx} 
\usepackage[inkscapelatex=false]{svg}
\usepackage{colortbl}
\usepackage{xcolor}
\usepackage{array}
\usepackage{caption}
\newcolumntype{Y}{>{\centering\arraybackslash}X}
\tcbset{enhanced,breakable} 
\definecolor{entityred}{RGB}{200, 38, 38}
\definecolor{infogold}{RGB}{230, 150, 0}
\definecolor{relblue}{RGB}{38, 90, 170}
\definecolor{relgreen}{RGB}{34, 139, 34}
\definecolor{deeppurple}{RGB}{85, 26, 139}
\definecolor{verycommon}{RGB}{34,139,34}   
\definecolor{common}{RGB}{255,165,0}      
\definecolor{uncommon}{RGB}{220,20,60}   
\usepackage{color}

\usepackage{balance} 

\title{KGHaluBench: A Knowledge Graph-Based Hallucination Benchmark for Evaluating the Breadth and Depth of LLM Knowledge}

\author{\textbf{Alex Robertson}$^{1}$ \quad \textbf{Huizhi Liang}$^{1}$ \quad \textbf{Mahbub Gani}$^{2}$ \\
  \textbf{Rohit Kumar}$^{2}$ \quad \textbf{Srijith Rajamohan}$^{2 \thanks{Current affiliation: Redis}}$ \\
  $^{1}$School of Computing, Newcastle University \quad $^{2}$Sage Ai, Sage Group PLC\\
  \texttt{\{a.robertson4, huizhi.liang\}@ncl.ac.uk} \\
  \texttt{\{mahbub.gani, rohit.kumar2, srijith.rajamohan\}@sage.com}
}

\begin{document}
\maketitle 
\begin{abstract}
Large Language Models (LLMs) possess a remarkable capacity to generate persuasive and intelligible language. However, coherence does not equate to truthfulness, as the responses often contain subtle hallucinations. Existing benchmarks are constrained by static, narrow questions, resulting in limited coverage and misleading evaluations. We present \textbf{KGHaluBench}, a \textbf{K}nowledge \textbf{G}raph-based \textbf{hal}l\textbf{u}cination \textbf{bench}mark that assesses LLMs across the breadth and depth of their knowledge, providing a fairer and more comprehensive insight into LLM truthfulness. 
Our framework utilises the KG to dynamically construct challenging, multifaceted questions, whose difficulty is then statistically estimated to address popularity bias.
Our automated verification pipeline detects abstentions and verifies the LLM's response at both conceptual and correctness levels to identify different types of hallucinations.
We evaluate 25 frontier models, using novel accuracy and hallucination metrics. The results provide a more interpretable insight into the knowledge factors that cause hallucinations across different model sizes.
KGHaluBench is publicly available\footnote{Code available at \url{https://github.com/c0037654Newcastle/KGHaluBench}} to support future developments in hallucination mitigation.
\end{abstract}


\section{Introduction}
Recent advancements in language generation and critical reasoning of Large Language Models (LLMs) have driven their widespread adoption \cite{llm_survey_2025}, with their remarkable capabilities revolutionising domains such as science, healthcare, and finance \cite{sci_llm_2023, health_llm_2025, fin_llm_2024}. However, a detrimental yet inevitable limitation with LLMs lies in their tendency to generate inaccurate or misleading content, often referred to as 'hallucinations' \cite{halu_inevitable_2024}. These may diverge from the intent of the user's input, contradict previously established outputs, and/or conflict with verifiable factual knowledge \cite{siren_song_2024}, undermining the LLM's trustworthiness, interpretability, and factual accuracy \cite{halu_survey_2024}. As more proficient LLMs generate increasingly convincing outputs, users could perceive these responses as the truth, amplifying the potential consequences of hallucinations. This emphasises the need for robust benchmarks to quantify hallucinations and assess the effectiveness of mitigation strategies.

%
The breadth of knowledge assessed in current benchmarks is often constrained by their static nature. This flaw often limits their ability to reflect realistic accuracy and hallucination rates. Some pre-constructed QA benchmarks capture the knowledge before and surrounding the time of their release \cite{webquestions-2013, simplequestions-2015, nq-2019}, but are unable to evaluate a model's performance on newly emerging facts and information. Incorporating a knowledge graph (KG) can alleviate these limitations; however, several KG-based benchmarks revert to static datasets for dynamic generation, thereby limiting the range of assessment and failing to reflect the diversity of knowledge users' queries. \cite{kg-fpq-2024, head-to-tail-2024, dynamic-kgqa-2025}. 

The depth of knowledge is rarely probed beyond surface-level details due to the question style used in the benchmarks. Often, closed or multiple-choice questions are favoured over open-ended questions \cite{MMLU_2021, DefAn_2024}, as accurately evaluating long-form responses is a current research challenge. However, this question style is knowledge-restrictive, limiting the LLM to choose from a specific set of answers. Instead, some benchmarks use simple questions \cite{simplequestions-2015, triviaqa-2017}, which are typically short, open-ended queries with a single, verifiable answer. These questions require the LLM to draw on its internalised representations, but are unable to capture multiple elements of deeper knowledge within a single question.

To address the limitations above, we present KGHaluBench, a \textbf{KG}-synergised \textbf{hal}l\textbf{u}cination \textbf{bench}mark. We leverage the comprehensive information within a KG to assess both factual accuracy and hallucination rate across an LLM's knowledge.
Our benchmark includes a dynamic question-generation module that draws on the KG to retrieve a random selection of entities spanning diverse topics. With an entity as the focal point, we generate an open-ended compound question that targets three aspects of information, both activating and assessing the depth of an LLM's knowledge.
Our response verification framework assesses the factuality of long-form text by identifying hallucinations in the LLM's output. The framework includes an abstention filter to detect expressions of uncertainty, an initial entity-level filter to identify semantic misalignment with the entity, and a final fact-level check to verify correctness against grounded facts.
We introduce novel metrics to evaluate our extensive experimental results, Weighted accuracy, $W_{\text{a}}$, scales standard accuracy based on the estimated difficulty of the content within the assessment, providing a fairer measure of performance. Breadth of knowledge and depth of knowledge hallucination rates, $\mathrm{Halu}_{\mathrm{BOK}}$ and $\mathrm{Halu}_{\mathrm{DOK}}$, are derived from what stage in the response verification framework the hallucination was detected, offering an insight into which aspects of the LLM's knowledge caused the hallucination.
Our key contributions are:
\begin{itemize}[left=0pt]
    \item We propose a question-generation approach that leverages the KG's relational structure to formulate compound questions over dynamically selected entities.
    \item We construct an automated verification framework with coarse entity-level and fine-grained fact-level filters to assess the factuality of the LLM's output, achieving 79.19\% and 87.74\% agreement with human judgment, respectively.
    \item We develop a statistical method to assess the difficulty of an assessment and scale the accuracy accordingly, generating a novel metric, $W_{\text{a}}$.
    \item We conduct an extensive experiment using 25 open-source and proprietary LLMs, leveraging $\mathrm{Halu}_{\mathrm{BOK}}$ and $\mathrm{Halu}_{\mathrm{DOK}}$ to identify factors in LLMs' knowledge that may cause hallucinations.
\end{itemize}

\section{Related Work}
\label{sec: Related_Work}
\paragraph{QA Benchmarks:}
The Question Answer (QA) benchmarks typically evaluate LLMs through open-ended generation, with results assessed by metrics or automated judges.
%
SimpleQA \cite{simpleqa-2024} utilises short, fact-seeking questions with a single, verifiable answer. While HotpotQA \cite{hotpotqa_2018} introduces additional complexity, requiring multi-hop reasoning to arrive at the answer.
These static QA benchmarks quickly become outdated, failing to challenge the constantly advancing LLMs. While dynamic QA benchmarks such as FreshQA \cite{freshqa-2023} and RealTimeQA \cite{realtimeqa-2023} exist, they are challenging to maintain due to complicated data management and frequent updates.
Knowledge-Graph Question-Answer (KGQA) benchmarks address this by using KGs, such as Wikidata and DBpedia \cite{wikidata-2014, dbpedia-2007}, to generate questions. 
While classic KGQA benchmarks such as ComplexWebQuestions \cite{cwq_2018} and FreebaseQA \cite{freebaseqa_2019} are static, using fixed KG snapshots, GraphEval \cite{grapheval-2024} and Dynamic-KGQA \cite{dynamic-kgqa-2025} dynamically construct test datasets to remain up-to-date.
However, random sampling from the KG may introduce an entity-popularity bias, leading to assessments dominated by well-known entities.
To mitigate this, some KGQA benchmarks incorporate discrete difficulty levels. KG-FPQ \cite{kg-fpq-2024} generates confusability levels based on the editing methods, whereas Head-To-Tail \cite{head-to-tail-2024} calibrates difficulty based on the entity popularity.
Yet discrete difficulties can create fairness issues when questions of varying difficulty are grouped. Therefore, KGHaluBench statistically estimates the difficulty of each question, aggregates for the assessment, and scales the accuracy accordingly, ensuring reliable evaluation regardless of the assessment's content.

\paragraph{Hallucination Benchmarks:}
Numerous benchmarks have been developed to assess an LLM's tendency to generate plausible but factually unsupported content. 
HalluLens \cite{hallulens_2025} separates hallucinations from fact by evaluating the LLM’s response to extrinsic and intrinsic hallucinations. TruthfulQA \cite{truthfulqa-2021} focuses on response truthfulness using questions replicating false beliefs or misconceptions. HaluEval \cite{halueval_2023} uses hallucinated samples to evaluate an LLM's ability to detect hallucinations.
However, many benchmarks rely on one-dimensional metrics, such as Accuracy, Accept/Refusal rates, BLEU, and BERTScore, which limit the interpretability of the results and leave the underlying causes of LLMs’ performance unclear.
To address this, benchmarks utilise more advanced verification approaches. FEVER \cite{fever_2018} uses a Natural Language Inference (NLI) model to evaluate if the response contains, contradicts, or does not mention the evidence.
FactCC \cite{factcc_2020} uses a BERT-based model to verify the factual overlap between a response and the evidence.
FactScore \cite{factscore_2023} decomposes a generation into atomic facts to calculate the amount which are supported by evidence. 
However, these approaches still collapse results into aggregate metrics that fail to provide clear, actionable insights.
KGHaluBench instead decomposes the common hallucination rate into $\mathrm{Halu}_{\mathrm{BOK}}$ and $\mathrm{Halu}_{\mathrm{DOK}}$ to determine the knowledge level responsible for the hallucination.

\begin{figure*}[!t]
    \centering
    \includegraphics[width=0.98\textwidth]{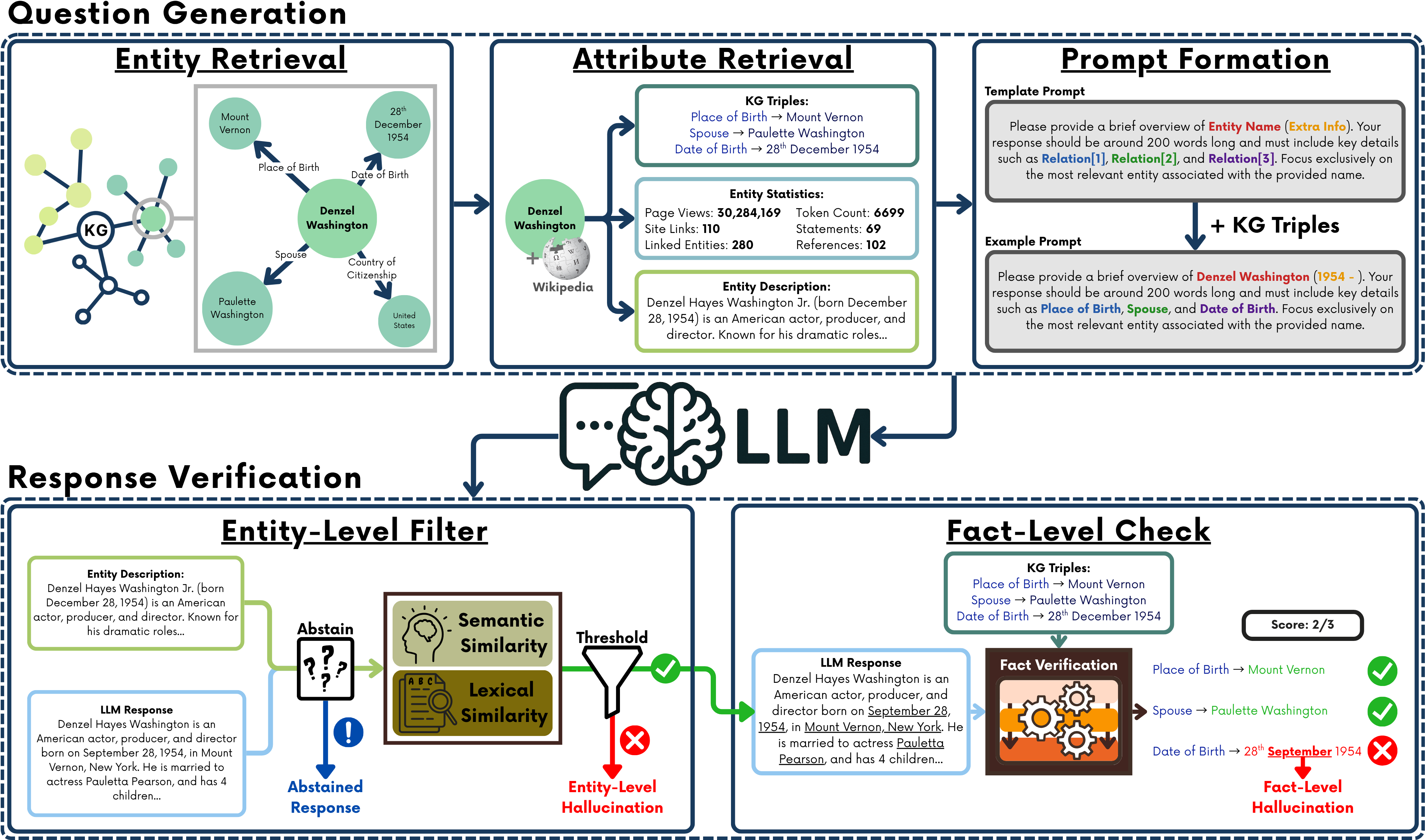}
    \caption{Framework of the KGHaluBench.}
    \label{fig: framework}
\end{figure*}

\section{The \textbf{KGHaluBench} Benchmark}
We propose KGHaluBench, a novel benchmark for evaluating the truthfulness of LLMs when answering challenging questions that span the breadth and depth of their knowledge. As shown in Figure \ref{fig: framework}, the benchmark consists of two complementary components. The \textit{Question Generation Module} extracts a random entity (\textit{e.g. Denzel Washington}) from the KG to become the centre of the question. The KG's structure, together with external databases, is then leveraged to fetch the entity's \textit{KG Triples, Statistics}, and \textit{Description} needed to construct and validate the multifaceted question. The \textit{Response Verification Module} employs a two-layer framework that first checks the \textit{LLM's response} against the entity's \textit{Wikipedia Description} to ensure non-abstention and confirm a basic understanding of the focal entity (\textit{e.g. Denzel Washington}). Non-hallucinated responses are then verified at the fact level by comparing their claims to the \textit{KG Triples} to ensure correctness.

\subsection{Question Generation}
The \textit{Question Generation Module} dynamically constructs coherent and challenging questions utilising the KG to alleviate the static, pre-constructed nature of current QA benchmarks.

\subsubsection{Entity Retrieval:}
\label{sec: ent_retrieval}
\begin{figure}[!htbp]
    \centering
    \includegraphics[width=0.475\textwidth]{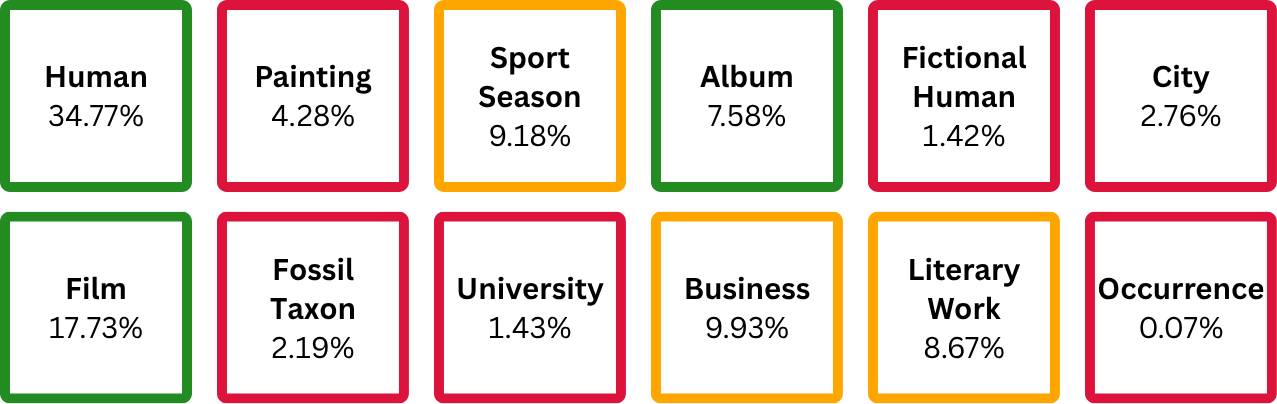}
    \caption{Distribution of \textcolor{verycommon}{Very Common}, \textcolor{common}{Common}, and \textcolor{uncommon}{Uncommon} Entity Types within the KGHaluBench assessment}
    \label{tab: Type_Distribution}
\end{figure}

For each question, we first define a focal entity. Using a batch call to the KG, we retrieve a random sample of entities, recording each ID and corresponding type. Invalid types are filtered against a predefined list of valid entity types, categorised by their KG frequency as \textcolor{verycommon}{Very Common} \textit{(e.g., Human)}, \textcolor{common}{Common} \textit{(e.g., Business)}, or \textcolor{uncommon}{Uncommon} \textit{(e.g., Painting)}. We order the sampling prioritising less common entities to maintain a balanced distribution of entity types, ensuring the benchmark remains topically broad yet structured. Table~\ref{tab: Type_Distribution} illustrates the resulting distribution of entity types in an average benchmark assessment. From the ordered sample, we sequentially select the focal entity for the question and extract its one-hop neighbours to form a subgraph. If no valid entity types are present in a sample, a new one is generated.

\subsubsection{Attributes Retrieval:}
\label{sec: att_retrieval}
We retrieve three key sets of information from the focal entity's sub-graph and external database for the downstream process in KGHaluBench.

\textit{KG Triples} are required both for generating the benchmark questions and verifying the correctness of the LLM's response. From the entity's subgraph, we extract pairs of facts and connecting relations and filter them against a predefined set of valid relations. As non-textual (e.g., \textit{image}), trivial (e.g., \textit{given name}), and irrelevant (e.g., \textit{official website}) relations are unsuitable for generating challenging and coherent questions. From the remaining pairs, we randomly select three to form the question and validate the response. If fewer than three valid pairs exist, we lack the number to formulate the question. Therefore, we discard the current target entity and select the next focal entity from the KG sample.

\textit{Entity Description} is a comprehensive overview of the entity from an external database. utilise the description as a factual representation when comparing against the LLM's response in the entity-level filter, as described in Section~\ref{Sec: res_verification}.

\textit{Entity Statistics} are used within the question difficulty calculation to estimate the popularity of the entity, as detailed in Section~\ref{sec: ent_popularity}. Similarly, in the KG-triple filtering process, if any statistic is null or invalid, we cannot quantify the entity's popularity; therefore, we must select the next focal entity from the sample set.
\subsubsection{Entity Popularity:}
\label{sec: ent_popularity}
Well-known entities are more likely to be referenced in an LLM's training data, increasing the likelihood that the LLM will accurately recall information about them. Therefore, estimating entity popularity provides valuable insight into the challenge posed by the question. To determine an entity's popularity, we combine the entity's individual relevance and the relevance of its associated type.
 
\paragraph{Entity Relevance:}
An entity's relevance is determined by two key factors: its \textcolor[HTML]{009999}{\textbf{prominence}}, reflecting recognisability and graph connectivity, and its \textcolor[HTML]{b66dff}{\textbf{information coverage}}, capturing the availability and detail of the information. To estimate relevance, we aggregate the following statistics from the KG and external databases. 

{\fontsize{8pt}{9pt}\selectfont
\begin{itemize}[leftmargin=1.2em, itemsep=3.5pt, parsep=0pt, topsep=4pt]
    \item \textcolor[HTML]{009999}{\textbf{Page Views}}: Number of entity page views (2017–2025)
    \item \textcolor[HTML]{009999}{\textbf{Site Links}}: Number of site links from other entities
    \item \textcolor[HTML]{009999}{\textbf{Linked Entities}}: Number of neighbours in the KG
    \item \textcolor[HTML]{b66dff}{\textbf{IDs}}: Number of external database identifiers
    \item \textcolor[HTML]{b66dff}{\textbf{Wiki Count}}: Token count of the entity's page
    \item \textcolor[HTML]{b66dff}{\textbf{Statements}}: Number of relation–fact pairs
    \item \textcolor[HTML]{b66dff}{\textbf{References}}: Number of sources validating the KG facts
\end{itemize}
}

However, some statistics provide a stronger indication of the LLM's ability to answer questions accurately. To account for this, we derive an associated weight for each statistic using a machine learning pipeline as outlined in Appendix~\ref{sec: weight_config}.

\paragraph{Entity Type Relevance:}
In the LLM's training data, some entity types (e.g., \textit{fictional characters}) frequently appear in descriptive texts, leading to a more comprehensive representation. In contrast, other entities (e.g. \textit{paintings}) are usually referenced in shorter, literal descriptions and are less likely to be retained. To quantify this difference, we calculate entity-type weights by averaging the question scores for each entity type and normalising them to the range [0, 1]. 

\subsubsection{Question Formation:}
\label{sec: prompt_form}
\begin{figure}[!htbp]
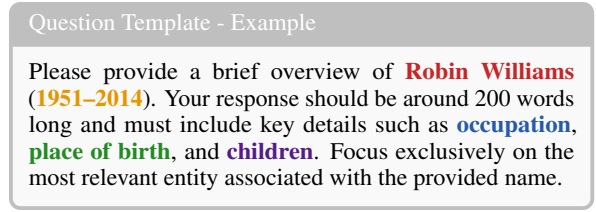

\centering
\begin{minipage}{1\linewidth}
\begin{tcolorbox}[
  colback=gray!5,
  colframe=gray!50,
  title=Question Template - Example,
  fonttitle=\footnotesize,
  fontupper=\footnotesize,
  boxsep=3pt, 
  left=3pt,          
  right=3pt,         
  top=3pt,            
  bottom=3pt          
]
Please provide a brief overview of \textcolor{entityred}{\textbf{Robin Williams}} (\textcolor{infogold}{\textbf{1951–2014}}). Your response should be around 200 words long and must include key details such as \textcolor{relblue}{\textbf{occupation}}, \textcolor{relgreen}{\textbf{place of birth}}, and \textcolor{deeppurple}{\textbf{children}}. Focus exclusively on the most relevant entity associated with the provided name.
\end{tcolorbox}
\end{minipage}
\caption{Question Template used for Question Generation.}
\label{fig: q_template}
\end{figure}

A key feature of KGHaluBench is the depth of its questions, which challenge the LLM's knowledge. We achieve this utilising the KG's relational structure to formulate compound questions about a single entity. As shown in Figure~\ref{fig: q_template}, our question template prompts the LLM to provide a brief overview of the entity and specific facts. This format is intentionally aligned with the entity's description to facilitate an accurate comparison during response verification. 

The template requires details such as the entity's name and three randomly chosen valid relations. We provide supplementary context about the entity to reduce ambiguity when multiple entities may share the same name, enabling the LLM to give a more accurate response without revealing any examined information. The completed question template is supplied to LLM, and the response is collected for verification. Implementation details are provided in Appendix~\ref{app: sup_context}.

\paragraph{Question Complexity:}
\label{sec: q_complexity}
To correctly answer each relation in the question, the LLM must recall specific facts; however, the availability of such knowledge can vary considerably. For instance, the album's artist is often better known and more widely documented than the record label that produced it.

To account for this, we adopt the same weight calculation process as used in Section~\ref{sec: ent_popularity}. As each entity type is associated with a distinct set of relations, we treat relation sets as independent. Within each set, we calculate the average question score for each valid relation and apply min-max normalisation to obtain relation weights. Question complexity is then estimated by aggregating the weights associated with the three relations it contains, reflecting the challenge the assessment content poses to the LLM. However, when the response is judged to be conceptually hallucinated, the complexity stems from the focal entity of the question rather than the relations. In this case, we approximate complexity by averaging the focal entity's relation set weights and multiplying the result by 3 to reflect the number of relations per question.

\subsubsection{Question Difficulty:}
\label{Sec: q_difficulty}
Dynamic question generation introduces randomness, leading to variations in difficulty across assessments. Therefore, we incorporate a difficulty-scaled accuracy metric to ensure a fair and consistent benchmark.

We create a unified measure of question difficulty, $Q_{\text{d}}$, by aggregating entity popularity and question complexity. We drew inspiration from Item Response Theory, a psychometric paradigm for test scoring. The theory employs a sigmoid function, which, due to its S-shaped curve, allows for greater differentiation in the middle and the most common range of difficulties. This ensures that moderate-difficulty questions contribute equally to those at the extremes, resulting in a fairer representation of all question difficulties. Our modified sigmoid formula is illustrated below, where \(\alpha\) controls the steepness of the sigmoid curve, \(Q\) is the question complexity, and \(Q_{\text{Avg}}\) represents the average question complexity across the three relations. \(EP\) denotes the entity popularity, which when min-max normalised is \(EP_{\text{Norm}}\)
\begin{eqnarray}
Q_{\text{d}} = \frac{1}{1 + e^{-\alpha \left(Q_{\text{Avg}} - EP_{\text{Norm}}\right)}}
\end{eqnarray}

\subsection{Response Verification}
\label{Sec: res_verification}
The \textit{Response Verification Module} addresses the challenge of evaluating long-form responses by accurately identifying abstentions and potential hallucinations, then verifying factual correctness. Each component of the framework was validated against human judgment to ensure accuracy and effectiveness, as detailed in Appendix~\ref{sec: res_verification}.

\subsubsection{Entity-Level Filter:}
\label{sec: ent_filter}
The entity-level filter classifies each response as aligned, hallucinated, or abstained, by identifying abstentions and evaluating the semantic and token-level similarities against the entity's description. Only aligned responses proceed to fact-level verification; hallucinated and abstained responses are scored accordingly and initiate another repetition of the question generation process.

We employ a small, efficient LLM to determine whether responses are abstentions or meaningful attempts to answer the question. The abstentions are defined as responses that refuse to answer, deflect the question, or admit to not recognising the focal entity. The models receive partial credit for appropriately expressing uncertainty or lack of knowledge~\cite{modelshallucinate_2025}, as we assign one point to abstained responses.

For responses that meaningfully address the question, we evaluate their alignment with the entity using both conceptual overlap and token-level intersection. For the semantic comparison, we encoded both the response and the description, then quantified their similarity using cosine similarity. For token-level comparison, we use the intersection of common words between the LLM's response and the entity's description. Entity-level similarity combines these metrics with a 70:30 ratio, prioritising semantic over lexical alignment, as the entity-level filter focuses on filtering out responses that misalign with the entity's domain. Responses exceeding the predefined threshold are considered aligned to the focal entity, while those below are deemed entity-level hallucinations and receive zero points. See Appendices~\ref{app: ent_equations} and~\ref{sec: res_verification} for further implementation details of the entity-level filter.

\subsubsection{Fact-Level Check:}
\label{sec: fact_filter}
\begin{figure}[!htbp]
    \centering
    \includegraphics[width=0.475\textwidth]{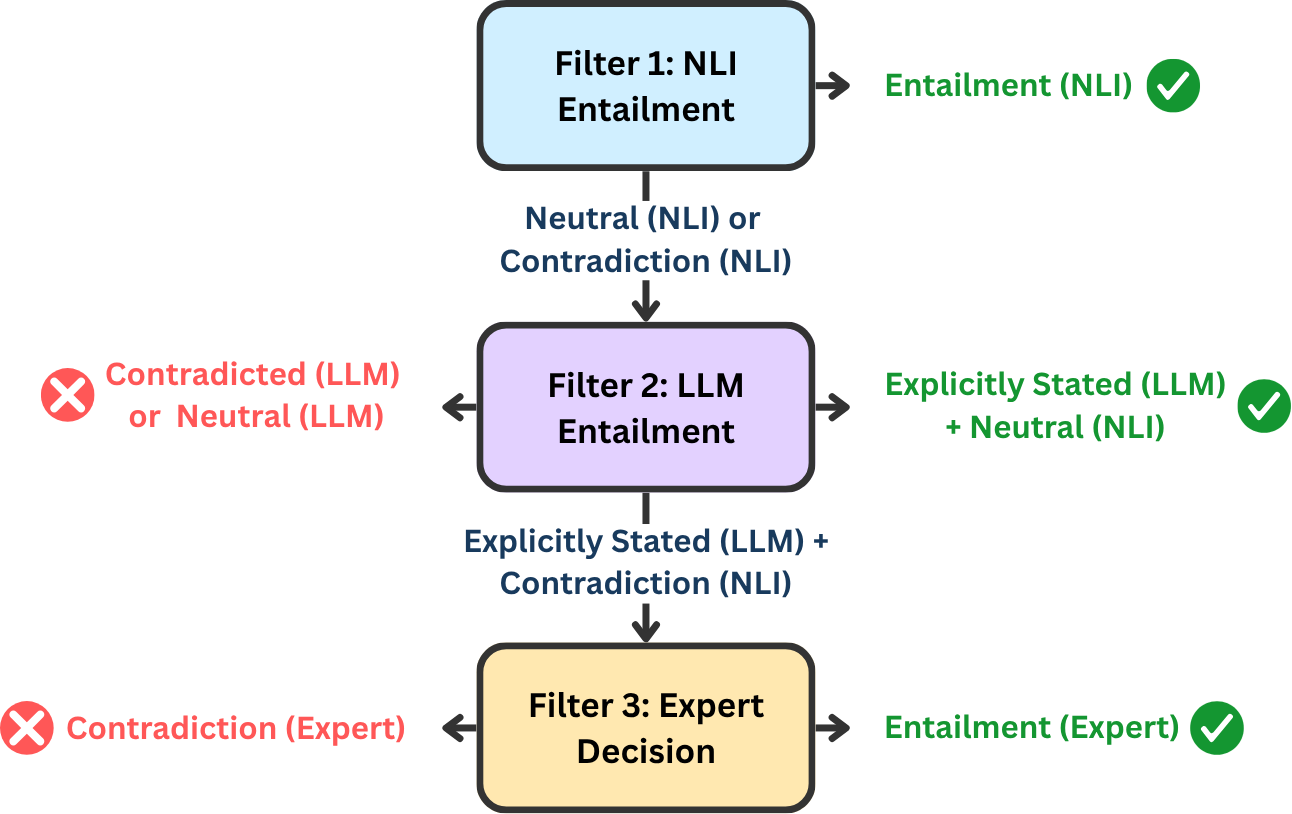}
    \caption{Overview of the fact verification pipeline}
    \label{fig: fact_pipeline}
\end{figure}

We implement a fact verification pipeline, as illustrated in Figure~\ref{fig: fact_pipeline}, to evaluate whether the relations specified in a question are correctly expressed in the LLM's response. Each relation is assessed independently as correct or incorrect, totalling a maximum of 3 points per response.

The pipeline requires an interpretable fact to compare to the LLM's response. Therefore, we employ an efficient LLM to transform the structured tuple of the entity's name, type, relation, tense indicator and fact into a sentence. The tense indicator is an auxiliary verb (e.g. 'is' or 'was') to ensure the constructed sentence is grammatically correct. 

\textit{NLI Entailment Filter:} For this filter, we utilise a Natural Language Inference (NLI) model. We chose this model because it is fast and efficient, while still achieving robust entailment recognition. We provided the NLI model with the reformatted fact and the LLM's response, requiring it to process the information and classify the outcome as entailment, contradiction, or neutral. If the result is entailment, we consider the fact correctly expressed in the LLM's response; otherwise, the response is passed to the LLM entailment filter.

\textit{LLM Entailment Filter:} We employ an LLM in place of an entailment model, exploiting its advanced reasoning abilities to make the filter more accurate; however, it is more computationally expensive. We prompt the LLM to assume the role of a fact-checking assistant, tasked with determining whether the following fact is explicitly stated, contradicted, or not mentioned in the provided response. If a fact is stated incorrectly or omitted from the response, it is considered incorrect. Whereas, if the LLM concludes that the fact is explicitly stated, its decision is cross-validated against the NLI model. If the NLI result is neutral, we accept the LLM's judgment and consider the fact correctly expressed in the response. However, if the NLI model contradicts the LLM's decision, we proceed to the expert decision filter.

\textit{Expert Decision Filter:} This filter functions as the final decision-maker. Within the pipeline, it is rarely used but serves as a fail-safe mechanism. We present the formatted fact alongside the response and create an ultimatum, with one expert's choice as entailment and the others as contradiction. We utilise the LLM to make the binary decision of which expert to agree with. If the LLM agrees with entailment, the fact is included in its response; otherwise, it is not. 

\section{Experiments}
\subsection{Knowledge Graph}
Although our framework supports integrating any KG, we selected Wikidata for the experiments. This expansive knowledge base, roughly 118 million entities as of August 2025 \cite{wiki_stats_2025}, enables the curation of novel, challenging benchmarks to assess the LLM's knowledge.

\subsection{Evaluation Models}
We assessed 25 state-of-the-art LLMs on the KGHaluBench benchmark, as detailed in Appendix~\ref{app: results_section}. This included 15 open-source models ranging from 8 billion to 1 trillion parameters, and 10 proprietary models from 4 leading AI companies: OpenAI, Google, Anthropic, and xAI. Each model was evaluated using default parameters across 10 runs of 150 questions, with results aggregated using the mean. Open-source models run on Nebius AI Studio API, while proprietary models were accessed via their respective APIs.

\subsection{Metrics} 

\subsubsection{Weighted Accuracy:}
We calculate $Accuracy$ as the percentage of possible points earned through correct facts and abstentions, rewarding models for both factuality and expressing uncertainty. However, to ensure we create a fair and consistent metric, $Accuracy$ must be scaled by the degree to which $Q_{\text{d}}$ deviates from the average difficulty across all assessments, $Avg(Q_{\text{d}})$. We denote the new metric as $W_{\text{a}}$.
\begin{eqnarray}
W_{\text{a}} = Accuracy \cdot \tfrac{Q_{\text{d}}}{Avg(Q_{\text{d}})}
\end{eqnarray}

\subsubsection{Hallucination Rate:}
We split the hallucination rate into breadth and depth of knowledge to provide more interpretability to an often one-dimensional metric.

$\mathrm{Halu}_{\mathrm{BOK}}$ is the percentage of responses classified as hallucinations by the entity-level filter, which assesses the surface-level truthfulness between the LLM's response and the entity's description. A high rate indicates that the LLM lacks fundamental knowledge of the assessment content, suggesting limited breadth of knowledge.
\begin{eqnarray}
\mathrm{Halu}_{\mathrm{BOK}} = 
\tfrac{|\text{Entity Hallucinations}|}{|\text{Total Responses| } - \text{ |Abstentions}|}
\end{eqnarray}
$\mathrm{Halu}_{\mathrm{DOK}}$ is the percentage of incorrect facts judged by the fact-level check. As only the responses that pass the entity-level filter are verified by the fact-level check, we divide the number of incorrect facts by the maximum possible score the model could achieve. A high rate indicates that while the LLM possesses the fundamental knowledge of the entity, it lacks sufficient depth to accurately answer the specified questions, suggesting a limited depth of knowledge.
\begin{eqnarray}
\mathrm{Halu}_{\mathrm{DOK}} =
  \tfrac{\text{|Incorrect Facts|}}{\text{Maximum Attainable Score}}
\end{eqnarray}

\section{Results}
\subsection{Weighted Accuracy}
\begin{figure}[!htbp]
    \centering
    \includegraphics[width=0.475\textwidth]{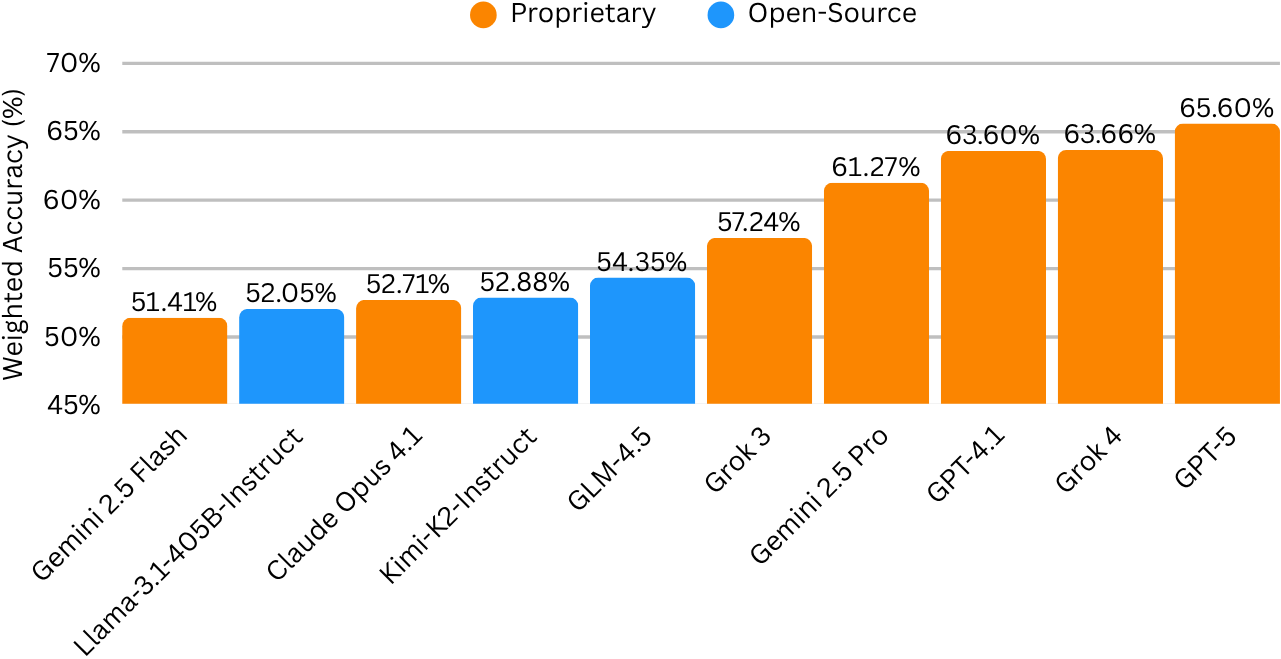}
    \caption{Top 10 Models by Weighted Accuracy. See Appendix~\ref{app: results_section} for the full table of results.}
    \label{fig: w_acc}
\end{figure}

KGHaluBench presents a substantial challenge for the LLMs, as displayed in Figure~\ref{fig: w_acc}, with \textit{GPT-5} achieving the highest $W_{\text{a}}$ of 65.60\%. This difficulty arises from the question-generation process, as the random focal entity around which the compound question is constructed may reside anywhere from the centre to the periphery of the LLM's knowledge. This redundant performance gap provides longevity for the KGHaluBench, enabling it to maintain its relevance against the continually advancing knowledge capabilities of LLMs.

The challenge of KGHaluBench makes it an effective benchmark for contrasting the factual performance between different LLMs. To enable a fair comparison between open-source and proprietary technologies, we compute the average $W_{\text{a}}$ for open-source models with $\geq 235$B parameters and compare it against the average performance of all proprietary models. The proprietary models demonstrate superior factuality, achieving an average $W_{\text{a}}$ of 55.94\%, compared to 48.32\%. This trend is evident: the five highest-performing models are all proprietary and show a clear increase in factuality compared to the rest. However, \textit{GLM-4.5} achieves 54.35\% $W_{\text{a}}$, outperforming powerful proprietary models, such as \textit{Claude-4-Opus} and \textit{Gemini-2.5-Flash}. This result indicates a narrowing performance gap between leading open-source and proprietary LLMs.

\subsection{Hallucination Rate}
Decomposing the hallucination rate provides valuable insight into the factors that contribute to hallucinations across different-sized models. To quantify these trends, we compare averages from the smaller 8-32B models with those from larger proprietary models. 
\begin{figure}[!htbp]
    \centering
    \includegraphics[width=0.48\textwidth]{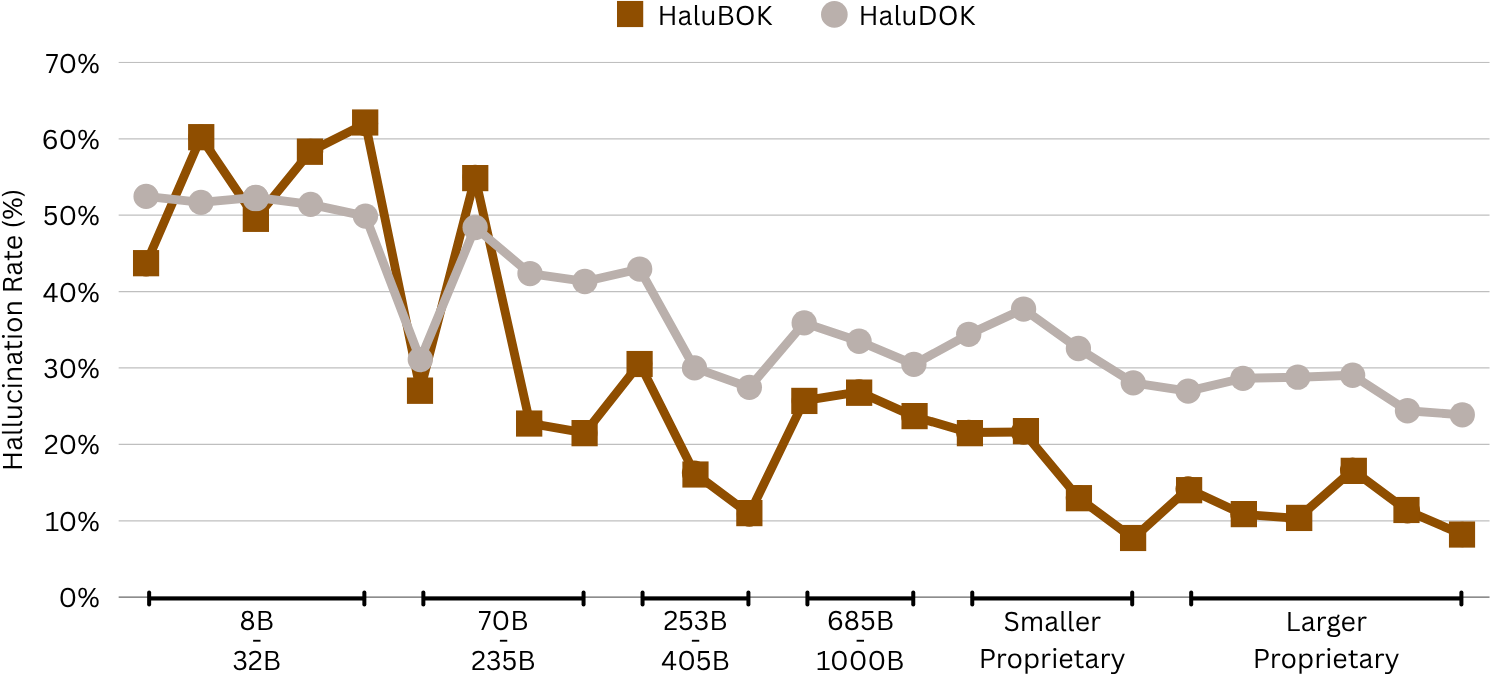}
    \caption{$\mathrm{Halu}_{\mathrm{BOK}}$ and $\mathrm{Halu}_{\mathrm{DOK}}$ hallucination rates across all models}
    \label{fig: hallu_rate}
\end{figure}
As shown in Figure~\ref{fig: hallu_rate}, the $\mathrm{Halu}_{\mathrm{BOK}}$ metric dramatically decreases, from 54.75\% to 11.91\%, as we move towards the larger, more proficient models. This trend suggests that the smaller models lack the competence to understand and recognise the focal entity in question. Therefore, the hallucinated responses are constructed from knowledge which is loosely or even unrelated to the entity. 

In contrast, the $\mathrm{Halu}_{\mathrm{DOK}}$ metric exhibits a more modest decrease by approximately 24.59\% from the 8-32B models to the larger proprietary models. This difference between the two metrics widens as the models become more proficient, suggesting that larger models excel at determining what to discuss but still struggle to retain the precise details of that topic. Therefore, the hallucinations are not due to a failure in entity recognition, but instead to a lack of deeper or more specific knowledge. 
\begin{figure}[!htbp]
    \centering
    \includegraphics[width=0.48\textwidth]{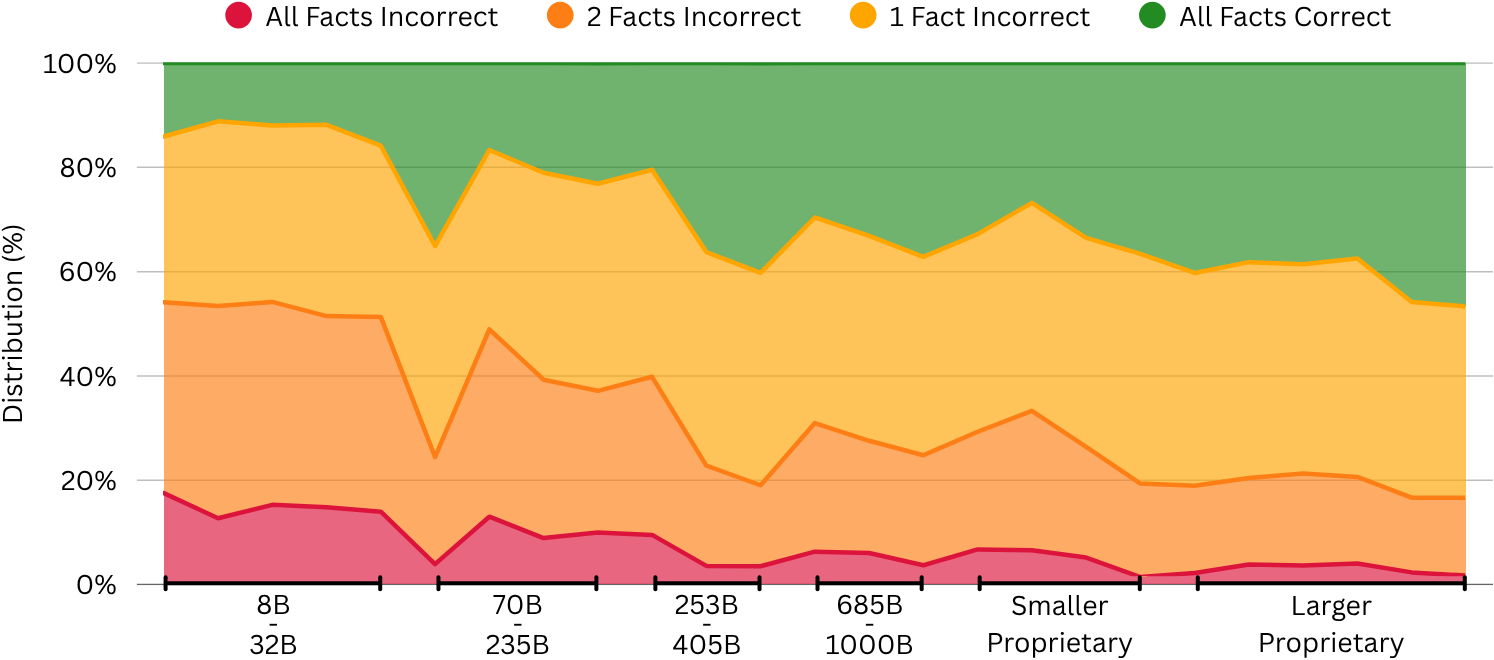}
    \caption{Distribution of fact-level hallucinations across all models, showing the proportion of responses containing 0, 1, 2, or 3 hallucinated facts}
    \label{fig: hallu_rate_dist}
\end{figure}

To further investigate this trend, we examine the fact-level hallucinations distribution across the models, as illustrated in Figure~\ref{fig: hallu_rate_dist}. The results show that more proficient models generate a lower proportion of highly detrimental responses, as those containing two or three hallucinated facts decrease from 52.91\% to 19.09\%. In turn, the percentage of responses that are entirely factually correct increases by around 28.18\%. However, roughly 39.74\% of responses from the more proficient models contain a single hallucinated fact. This undermines trustworthiness and creates a false sense of reliability, as the responses which appear fluent and factual still often include hallucinations.

\subsection{Abstention Rate}
\begin{figure}[!htbp]
    \centering
    \includegraphics[width=0.475\textwidth]{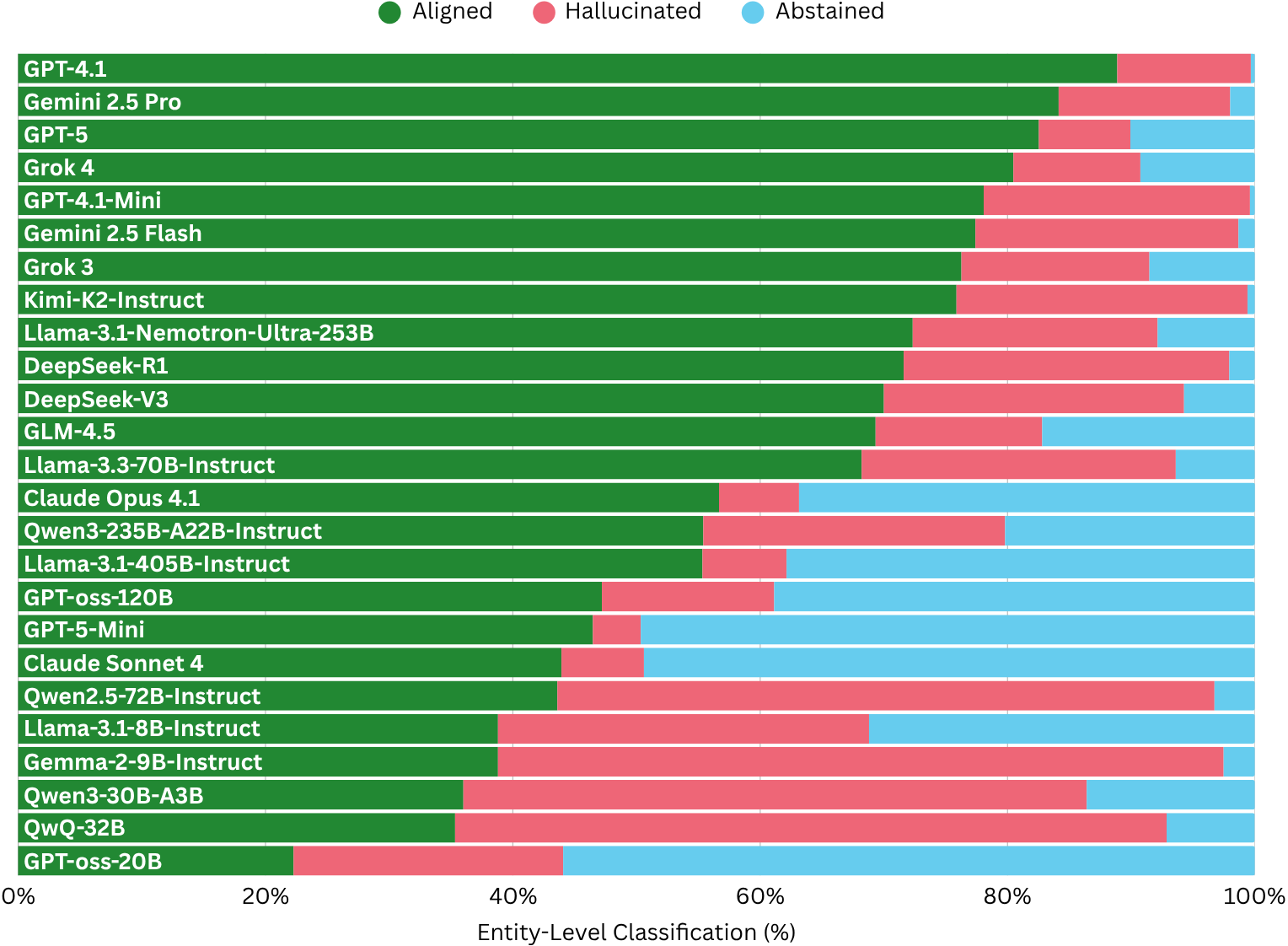}
    \caption{Distribution of entity-level filter classifications across all models, showing the proportion of responses classified as aligned, hallucinated, or abstained}
    \label{fig: abstain_fig}
\end{figure}

Abstention is an expression of uncertainty, where a model refrains from answering the question if it lacks the understanding or knowledge to provide an accurate response. We integrate abstention detection in our entity-level filter, detailed in Section~\ref{sec: ent_filter}, enabling the classification of each response as aligned, hallucinated, or abstained. Figure~\ref{fig: abstain_fig} reveals that among models with comparable alignment rates, those with higher abstention show lower hallucination rates. For example, \textit{Llama-3.1-8B} and \textit{Gemma-2-9B} both achieve a 38.80\% alignment between their responses and the focal entities' descriptions, indicating an equivalent breadth of knowledge. However, \textit{Llama-3.1-8B} has a higher abstention rate, resulting in a significantly lower hallucination rate of 30.00\%, compared to \textit{Gemma-2-9B} of 58.67\%. While this mechanism clearly reduces hallucinations, over-abstaining can severely limit the usefulness of the model. Several models demonstrate this problem with \textit{GPT-oss-20B}, \textit{GPT-5-Mini}, and \textit{Claude Sonnet 4}, declining to answer 55.93\%, 49.67\%, and 49.40\% of the benchmark questions, respectively. Therefore, we believe models should implement constructive abstention, in which the response provides guidance on how users can locate reliable information themselves.

\section{Conclusion}
We introduce KGHaluBench, a comprehensive hallucination benchmark that evaluates LLMs with questions that probe their depth and breadth of knowledge. Customisable framework dynamically generates multifaceted compound questions from randomly selected entities in the KG. These questions are then assessed at a conceptual level, and if no hallucinations or abstentions are detected, they are verified for factual correctness. Our experiments assess 27 state-of-the-art LLMs, utilising our novel metrics. We propose $W_{\text{a}}$, a fairer accuracy metric which adjusts standard accuracy by statistically estimating question difficulty. We also present $\mathrm{Halu}_{\mathrm{BOK}}$ and $\mathrm{Halu}_{\mathrm{DOK}}$, as more interpretable hallucination metrics to provide deeper insights into factors that lead different-sized LLMs to hallucinate. We hope KGHaluBench can be used to develop effective strategies in addressing the challenge of mitigating hallucinations.

\section{Limitations}
In our benchmark, we rely on Wikidata as the information source, limiting the set of fact-seeking questions to entities within the KG. 
This introduces data quality limitations, as Wikidata's representation of entities is uneven across topics, cultures, and languages, particularly under-representing non-English entities and marginalised topics.
Therefore, we would like to expand the benchmark further to include domain-specific KGs to explore hallucinations in specific scenarios. Additionally, our automated fact verification framework will make mistakes, potentially rejecting valid responses or scoring misaligned ones. Therefore, despite achieving substantial agreement with human judgment, as displayed in ~\ref{sec: res_verification}, we must make improvements in accuracy and soundness of reasoning, which would further strengthen the validity and reliability of KGHaluBench. 

\bibliography{references}


\appendix
\section{Appendix}
\subsection{Experimental Results}
\label{app: results_section}
Table~\ref{tab: results_table} details the weighted accuracy, abstain rate, and both hallucination rates for all models tested within the KGHaluBench experiments.

\begin{table*}[b]
\centering
\footnotesize
\begin{tabular*}{\textwidth}{@{\extracolsep{\fill}}p{4.75cm}cccccc}
\toprule
\textbf{Model Version} & Knowledge Cut-off &Model Size & $W_{\text{a}}$ & Abstain Rate & $\mathrm{Halu}_{\mathrm{BOK}}$ & $\mathrm{Halu}_{\mathrm{DOK}}$ \\
\midrule
\textit{Open-Source Models} & (Date) & (Billion) & (\%) & (\%) & (\%) & (\%) \\
\midrule
    Meta-Llama-3.1-8B-Instruct-fast & 12/2023 & 8 & 28.89 & 31.20 & 43.64 & 52.46 \\
    gemma-2-9b-it-fast & - & 9 & 19.54 & 2.53 & 60.21 & 51.67 \\
    GPT-oss-20b & 06/2024 & 20 & 29.45 & 55.93 & 49.56 & 52.32 \\
    Qwen3-30B-A3B & - & 30 & 22.20 & 13.60 & 58.36 & 51.41 \\
    QwQ-32B-fast & 11/2024 & 32 & 20.18 & 7.13 & 61.99 & 49.88 \\
    Llama-3.3-70B-Instruct-fast & 12/2023 & 70 & 49.04 & 6.40 & 27.13 & 31.10 \\
    Qwen2.5-72B-Instruct & 12/2023 & 72 & 23.67 & 3.27 & 54.96 & 48.39 \\
    GPT-oss-120b & 06/2024 & 120 & 40.62 & 38.87 & 22.74 & 42.36 \\
    Qwen3-235B-A22B-Instruct-2507 & - & 235 & 44.94 & 7.87 & 21.49 & 41.32 \\
    Llama-3\_1-Nemotron-Ultra-253B-v1 & 12/2023 & 253 & 38.61 & 20.20 & 30.57 & 42.95 \\ 
    GLM-4.5 & - & 335 & \textbf{54.35} & 17.20 & \underline{16.24} & \underline{30.01} \\
    Meta-Llama-3.1-405B-Instruct & 12/2023 & 405 & 52.06 & 37.87 & \textbf{10.88} & \textbf{27.46} \\
    DeepSeek-V3-0324 & 07/2024 & 685 & 47.00 & 5.73 & 25.72 & 35.90 \\
    DeepSeek-R1-0528 & 01/2025 & 685 & 48.45 & 2.07 & 26.87 & 33.50 \\
    Kimi-K2-Instruct & - & 1000 & \underline{52.88} & 0.60 & 23.66 & 30.47 \\
\midrule
\multicolumn{4}{l}{\textit{Proprietary Models}} \\
\midrule
    GPT-4.1-mini-2025-04-14 & 06/2024 & - & 48.13 & 0.40 & 21.62 & 37.71 \\
    GPT-4.1-2025-04-14 & 06/2024 & - & 63.60 & 0.33 & 10.84 & 28.65 \\
    GPT-5-mini-2025-08-7 & 06/2024 & - & 49.58 & 49.67 & \textbf{7.69} & 28.05 \\
    GPT-5-2025-08-07 & 10/2024 & - & \textbf{65.60} & 10.07 & \underline{8.20} & \textbf{23.87} \\
    Gemini-2.5-flash & 01/2025 & - & 51.41 & 1.33 & 21.55 & 34.41 \\
    Gemini-2.5-pro & 01/2025 & - & 61.27 & 2.00 & 14.18 & 26.97 \\
    Claude-sonnet-4-20250514 & 03/2025 & - & 46.18 & 49.40 & 13.01 & 32.55 \\
    Claude-opus-4-1-20250805 & 03/2025 & - & 52.71 & 36.87 & 10.32 & 28.79 \\
    Grok-3 & 11/2024 & - & 57.24 & 8.53 & 15.20 & 29.05 \\
    Grok-4-0709 & 11/2024 & - & \underline{63.66} & 9.27 & 10.27 & \underline{24.40} \\
\bottomrule
\end{tabular*}
\caption{Performance of 25 language models on the KGHaluBench using the metrics of $W_{\text{a}}$, $\mathrm{Halu}_{\mathrm{BOK}}$, and $\mathrm{Halu}_{\mathrm{DOK}}$, along with their version specifications. All models were evaluated across 10 assessments, each comprising 150 questions, with results aggregated using the mean. Models were evaluated using default parameters in accordance with their respective usage policies. The best performing results for the open-source and proprietary models are in \textbf{bold}, with the second-best values \underline{underlined}. The knowledge cut-off dates were collected from \url{https://github.com/HaoooWang/llm-knowledge-cutoff-dates}. Citations: Llama families \cite{llama3, llama-nemotron}, Qwen families \cite{qwen2.5, qwen3}, Gemma2\cite{gemma2}, GLM-4.5 \cite{GLM_4.5}, Kimi-K2 \cite{kimi_k2} DeepSeek families \cite{deepseek-r1, deepseek-v3}, GPT families \cite{gpt-4}, Gemini families \cite{gemini_2.5}}
\label{tab: results_table}
\end{table*}

\subsection{Detailed Methodology}
\subsubsection{Supplementary Context in the Question Template}
\label{app: sup_context}
The supplementary context provided in the question varies depending on the type of the focal entity. If the entity type is a human, the supplementary context is their lifespan: (1934-2023) if deceased, or their birth year to the present (1968-). For all other types, the context would be the entity type.

\subsubsection{Detailed Entity-Level Filter}
\label{app: ent_equations}
For the semantic comparison, we embedded both the response and the description into 1024-dimensional vectors using Jina-Embedding-V3~\citep{jina_2024}. The model achieved a Sentence Textual Similarity (STS) score of 85.80 on the MTEB benchmark while maintaining an efficient parameter count of 570 million, making it an ideal choice for entity-level filtering~\citep{mteb_2022}. We then quantified the similarity between the two texts using cosine similarity.

{\footnotesize
\begin{flushleft}
\begin{eqnarray}
\mathbf{w}=\text{Embed}(\text{Description}), \quad \nonumber \mathbf{r}=\text{Embed}(\text{Response}) \nonumber \\
\text{} \nonumber \\
\text{SemanticSim}=\dfrac{\mathbf{w} \cdot \mathbf{r}}{\|\mathbf{w}\| \|\mathbf{r}\|}
\end{eqnarray}
\end{flushleft}
}

We calculate the token-level comparison using the Fuzzy Set Ratio from the RapidFuzz Module, which is based on the Levenshtein Distance Formula~\citep{levenshtein_1966}. The Fuzzy Set Ratio uses the intersection of common words between the LLM's response and the entity's description, which is optimal for texts which differ in order and length, as is common in this situation. The Fuzzy Set Ratio, like cosine similarity, produces a similarity score represented as a percentage.

{\footnotesize
\begin{flushleft}
\begin{equation}
\begin{array}{l}
TokenSim = \left(1 - \frac{D(\text{Description}, \text{Response})}{\max(\lvert \text{Description} \rvert, \lvert \text{Response} \rvert)} \right)
\end{array}
\end{equation}
\end{flushleft}
}

We aggregate entity similarity with a bias toward semantic meaning, as it better captures the conceptual relationship between the two texts.

{\footnotesize
\begin{flushleft}
\begin{equation}
EntitySim = 0.7 \cdot SemanticSim + 0.3 \cdot TokenSim 
\end{equation}
\end{flushleft}
}

\subsubsection{Weights Configuration}
\label{sec: weight_config}
We created a calibration dataset by evaluating 34 selected models (1B-685B parameters, including both open-source and proprietary models) across 10 runs of 150 questions, resulting in a total of 51,000 data points. For each question, the dataset captures the corresponding entity's ID, statistics, and type, as well as the question's relation types, their scores, and the overall question score.

To ensure a balanced representation of all models in the weight configuration, we used stratified sampling to split the experiment dataset into training and validation sets at a 70:30 ratio. For each model's 1,500-question section, 70$\%$ (1,050 questions) were randomly selected for the training set, and the remaining 30$\%$ (450 questions) were reserved for validation.

To estimate the entity relevance weights, we employed TPOT to automatically generate a machine learning pipeline optimised to predict how well-known an entity is to the LLM, based on the entity's statistics. Before the training, we normalised each statistic using a natural logarithm transformation to ensure consistent scaling across features. The final pipeline includes max-abs feature scaling, feature selection using an ExtraTreesClassifier, parallel feature transformations via FeatureUnion, and a final ensemble using a BaggingClassifier. All hyperparameters and component choices were selected using TPOT's approach, which balances accuracy and model complexity. Finally, the learned feature coefficients were extracted and interpreted as entity-relevance weights.

Entity type relevance weights and relation complexity weights were also derived from the training data, following the processes outlined in Section~\ref{sec: ent_popularity} and Section~\ref{sec: q_complexity}, respectively.

\subsection{Ablation Studies}
We conduct an ablation study to isolate the contributions of the response verification framework and the weighted accuracy calculation. We demonstrate that both stages of the verification process closely align with human judgment. Additionally, we demonstrate that incorporating difficulty-based scaling into the accuracy metric results in a more consistent and fair assessment.

\subsubsection{Response Verification Framework}
\label{sec: res_verification}
We conducted a human validation study to evaluate the accuracy of the verification framework in comparison to human judgment. We tasked nine participants with answering 900 questions: 495 emulating the role of the entity-level filter and 405 replicating the role of the fact-level checker. For the entity-level filtering task, participants compared the LLM's response with the entity's Wikipedia description to determine if they referred to the same entity, disregarding any fact-level hallucinations. For the fact-level filtering task, participants were given three facts corresponding to the relations in the question and had to verify whether each was explicitly stated in the LLM's response. We compared our framework against an automated judge which utilised \textit{GPT-3.5-Turbo}, as often used in literature. The nine participants, all Master's or PhD students, provided consent and were made aware that their responses would be used to evaluate and validate the verification framework.

\paragraph{Entity-Level Filter:}
Table~\ref{tab: entity_filter_comparison} illustrates how varying the threshold influences the balance between precision and recall in the entity-level filter. We chose the specific thresholds after observing several KGHaluBench assessments to estimate the boundary between factual and hallucinated responses. The highest threshold of 0.750 achieved the greatest alignment with the human judges, at 79.19\%. However, the lower recall of 73.17\% indicates that the filter is overly strict for initial conceptual overlap comparisons, resulting in valid responses being wrongly rejected. Lowering the threshold increases recall faster than it decreases precision, resulting in the highest F1 score of 78.07\% at a threshold of 0.700, despite the overall agreement dropping slightly to 77.98\%. Since the entity-level filter is at the first stage of the pipeline, we prioritise recall, as misaligned responses admitted at this stage will score poorly at the fact-level check, whereas aligned responses mistakenly discarded are detrimental to the assessment's accuracy score. These results suggest that the entity-level filter effectively differentiated between conceptually hallucinated and non-hallucinated responses, in agreement with the participants' decisions. 

\begin{table}[!htbp]
    \footnotesize
    \centering
    \begin{tabular*}{0.48\textwidth}{@{\extracolsep{\fill}}llcccc}
        \toprule
        \textbf{Model} & & \textbf{P \%} & \textbf{R \%} & \textbf{F1 \%} & \textbf{A \%} \\
        \midrule
        \multirow{3}{*}{\textit{KGHaluBench}}
          & \textbf{0.750} & 75.76 & 73.17 & 74.44 & 79.19 \\
          & \textbf{0.725} & 70.45 & 84.88 & 76.99 & 78.99 \\
          & \textbf{0.700} & 66.44 & 94.63 & 78.07 & 77.98 \\
        \midrule
        \textit{GPT-3.5-Turbo} & \textbf{--} & 82.46 & 45.85 & 58.93 & 73.54 \\
        \bottomrule
    \end{tabular*}
    \caption{Entity-Level Filter Performance Across Thresholds Compared to \textit{GPT-3.5-Turbo} (P = Precision, R = Recall, A = Human Agreement)} 
    \label{tab: entity_filter_comparison}
\end{table}

We compared the entity-level filter against an automated judge using a backbone of \textit{GPT-3.5-Turbo}. Our 0.700 threshold achieved 5.65\% higher alignment with human judgment and 48.78\% higher recall. These results suggest that the entity-level filter is a far more effective approach than a typical automated judge, as it can reliably determine whether the focal entity is consistent across the two texts despite potential nuances and discrepancies.

\paragraph{Fact-Level Check:}
In the human validation study, our tri-stage fact verification pipeline achieved 87.74\% alignment with human judgment, which was 8.56\% higher than the automated judge at 79.18\%. To conduct a further investigation into the pipeline, we utilise the experimental data to highlight the contribution and efficiency of each filter.

\begin{table}[!htbp]
    \footnotesize
    \centering
    \begin{tabular*}{0.475\textwidth}{@{\extracolsep{\fill}}ccc}
        \toprule
        \textbf{Filter} & \textbf{Facts Verified (\%)} & \textbf{Avg Time (s)} \\
        \midrule
            NLI Entailment & 27.58 & 0.36 \\
            LLM Entailment & 71.39 & 1.91 \\
            Expert Decision & 1.03 & 2.35 \\
        \midrule
            \multicolumn{3}{c}{\textbf{Average Verification Time per Fact (s):} 1.49} \\
        \midrule
            \multicolumn{3}{c}{\textbf{Average Verification Time per Question (s):} 4.47} \\
        \bottomrule
    \end{tabular*}
    \caption{Distribution of verified facts and average processing time in the fact verification pipeline}
    \label{tab: fact_filter_table}
\end{table}

The first filter of the pipeline employs \textit{RoBERTa-Large-MNLI} as an efficient preliminary check. This stage contributes most to the pipeline's overall efficiency, resolving 27.58\% of the facts with an average verification time of 0.36 seconds, as shown in Table~\ref{tab: fact_filter_table}. The model's effectiveness is due to its lightweight design, with just 365M parameters. However, \textit{RoBERTa-Large-MNLI} has limited reasoning capabilities, which necessitates a second layer of verification. Therefore, we position the filter at the forefront of the pipeline, where it primarily encounters facts that are easily verifiable, such as explicitly stated or numerical ones.

The secondary check is managed by the LLM entailment filter, which utilises the semantic reasoning capabilities of \textit{Llama3.1:8B}. This stage contributes most to the pipeline's overall accuracy, as it handles cases with subtle discrepancies between the response and the grounded fact, such as changes in word order, phrasing, or nuance. Although roughly five times slower than the first filter, it still resolves 71.39\% of the facts, indicating its importance in maintaining the integrity of the fact verification pipeline.

The Expert Decision filter resolves only 1.03\% of the facts that pass through the fact verification pipeline. This outcome is expected, as a fact must meet a particular set of conditions in the first two filters before reaching this final fail-safe stage. Overall, the fact verification pipeline is efficient and accurate, which contributes to the validity of the results from KGHaluBench.

\subsubsection{Question Difficulty Weighting}
We utilise the training section of our calibration dataset, as detailed in Appendix~\ref{sec: weight_config}, to configure the weights used in the question difficulty calculation. To validate this metric, we assess how well the estimated difficulty aligns with the LLM's experienced difficulty by using the validation portion of the dataset. We use the question score to represent the actual difficulty, with higher scores indicating more straightforward questions. We then compared actual difficulty with estimated question difficulty and evaluated the relationship using Spearman's and Kendall's rank correlation coefficients to provide robust ordinal association measures.

The results show a moderate negative correlation between actual and estimated question difficulty, with Spearman's $\rho$ = -0.403 and Kendall's $\tau$ = -0.299. Kendall's $\tau$ is marginally lower, as it only considers the number of pairwise disagreements, whereas Spearman's $\rho$ also accounts for the distance between the pairs. This confirms that our metric captures the difficulty of the content within the benchmark questions.

To explore the practical impact of this metric, we apply the derived weights to calculate question difficulty scores for our main experimental evaluation. We assess whether it can improve assessment fairness by addressing the randomness and entity-popularity bias introduced by the question-generation mechanism. For each of the 25 models, we compute the mean accuracy and the weighted accuracy ($W_{\text{a}}$), along with their standard deviations, across 10 runs. These values are then averaged across all models to obtain aggregated metrics.

\begin{table}[!htbp]
    \footnotesize
    \centering
    \begin{tabular*}{0.40\textwidth}{@{\extracolsep{\fill}}ccc}
        \toprule
        \textbf{Metric} & \textbf{Mean Acc. (\%)}& \textbf{Std. Dev. (\%)}\\
        \midrule
            Accuracy & 45.30 & 2.57 \\
            $W_{\text{a}}$ & 45.25 & 2.45 \\
        \midrule
             & -0.05 & -0.12 \\
        \bottomrule
    \end{tabular*}
    \caption{Mean and Standard deviation of Accuracy and $W_{\text{a}}$, over the 10 runs per model. Then averaged across all models in the experiment dataset}
    \label{tab: acc_stdev_table}
\end{table}

As shown in Table~\ref{tab: acc_stdev_table}, applying difficulty-based weighting reduces the mean standard deviation across models by 0.12\%, from 2.57\% to 2.45\%. This improvement in consistency demonstrates that $W_{\text{a}}$ mitigates entity-popularity bias across assessments, resulting in a fairer, more stable benchmark. Notably, the weighting has a negligible effect on the mean accuracy, decreasing it by 0.05\% to 45.25\%. This ensures that $W_{\text{a}}$ continues to reflect the benchmark's intended difficulty, preserving the validity of the accuracy results.

\subsection{Implementation Details}
\tcbset{enhanced, breakable}

\subsubsection{Benchmark Prompts}

\begin{tcolorbox}[title=System Prompt for all LLMs, colback=gray!5, colframe=gray!95, fontupper=\ttfamily\scriptsize]
Answer questions about specific entities. If you recognise the focal entity, give a clear, detailed response covering all requested details. If you do not know it well enough, state that you cannot answer.
\end{tcolorbox}

\begin{tcolorbox}[title=Benchmark Template for all LLMs, colback=gray!5, colframe=gray!95, fontupper=\ttfamily\scriptsize]
Please provide a brief overview of [Entity Name] [Extra Context]. Your response should be around 200 words long and must include key details such as [Entity Relation 1], [Entity Relation 2], and [Entity Relation 3]. Focus exclusively on the most relevant entity associated with the provided name.
\end{tcolorbox}
Parameters: Given the model's default.


\subsubsection{Response Verification Prompts}
\paragraph{Entity-Level Prompts:}
Prompt used within the entity-level filter.

\begin{tcolorbox}[
  title=Abstention Filter (Llama3.2:3b),
  colback=gray!5,
  colframe=gray!95,
  width=\linewidth,
  fontupper=\ttfamily\scriptsize,
  before upper={\setlength{\parskip}{0.7\baselineskip}\setlength{\parindent}{0pt}}
]
You are given a question and a response from a language model. Determine if the response abstains from answering the question. A response counts as abstaining if it refuses, deflects, or avoids providing the requested information, or if it only gives generic disclaimers without including the required details. If the response meaningfully answers the question, classify it as 'Answered'.
 
Return only one word: 'Abstained' or 'Answered'.
 
\#\#\# Question: [Benchmark Question]

\#\#\# Response: [LLM Response]
\end{tcolorbox}
Parameters: Temperature: 0, Top\_P: 0.6

\paragraph{Fact-Level Prompts:}
Prompt used within the fact verification pipeline.

\begin{tcolorbox}[title=Tuple to Golden Fact Translation, colback=gray!5, colframe=gray!95, fontupper=\ttfamily\scriptsize]
Convert the following structured tuple into a natural language sentence, using the [Entity Name] as the subject: ([entity Name] ([Entity Type]) , [Entity Relation] , [Verb] , [Fact]). Return only the resulting sentence and do not reword [Entity Name] or [Fact].
\end{tcolorbox}
Parameters: Temperature: 0.3, Top\_P: 0.5 

\begin{tcolorbox}[
  title=LLM Entailment Filter (Llama3.1:8b),
  colback=gray!5,
  colframe=gray!95,
  width=\linewidth,
  fontupper=\ttfamily\scriptsize,
  before upper={\setlength{\parskip}{0.7\baselineskip}\setlength{\parindent}{0pt}}
]
You are a fact-checking assistant. Your task is to determine whether the following fact is explicitly stated, supported, contradicted, or not mentioned in the provided response.

\#\#\# Response:
[LLM Response]

\#\#\# Fact:
[Golden Fact]

\#\#\# Response Options:
Respond with one of the following options and a brief explanation:

- EXPLICITLY STATED: The fact is directly and clearly stated in the response, using the same or equivalent wording. Numerical or time-related facts must match exactly.
- CONTRADICTED: The fact is directly contradicted by information in the response.
- NOT MENTIONED: The fact is not present in the response, and there is no sufficient evidence to confirm or contradict it.

Only return one of the four options and a single concise explanation. Do not provide additional commentary.
\end{tcolorbox}
Parameters: Temperature: 0, Top\_P: 0.6

\begin{tcolorbox}[
  title=Expert Entailment Filter (Mistral:7b),
  colback=gray!5,
  colframe=gray!95,
  width=\linewidth,
  fontupper=\ttfamily\scriptsize,
  before upper={\setlength{\parskip}{0.7\baselineskip}\setlength{\parindent}{0pt}}
]
You are a fact-checking assistant. Your task is to determine whether Expert 1 or Expert 2 is correct based on the provided response and fact.

\#\#\# Response:
[LLM Response]

\#\#\# Fact:
[Golden Fact]

Expert 1: Entailment (The response aligns with the fact.)
Expert 2: Contradiction (The response contradicts the fact.)

Return only "Expert 1" or "Expert 2" based on the correct evaluation. No explanations.
\end{tcolorbox}
Parameters: Temperature: 0, Top\_P: 0.6


\subsubsection{Automated Judge Prompts}
Prompts used in the human validation study for comparing KGHaluBench's entity-level and fact-level filters against using an automated judge of GPT-3.5-Turbo.
\begin{tcolorbox}[
  title=GPT Entity-Level Filter,
  colback=gray!5,
  colframe=gray!95,
  width=\linewidth,
  fontupper=\ttfamily\scriptsize,
  before upper={\setlength{\parskip}{0.7\baselineskip}\setlength{\parindent}{0pt}}
]
You will be given two passages:
- A response generated by a large language model.
- A reference description of a target entity.

Your task is to determine if both passages refer to the **same entity**.

Check both:
1. Semantic similarity (do they describe the same person/place/thing, even with different wording?)
2. Token-level match (e.g., name or identifiers must match exactly or with minor variation)

Respond with only one word: True if they refer to the same entity, or False if they do not.

\#\#\# Response:
[LLM Response]

\#\#\# Description:
[Entity Description]

Output:
\end{tcolorbox}
Parameters: Temperature: 0, Max Tokens: 10

\begin{tcolorbox}[
  title=GPT Fact-Level Check,
  colback=gray!5,
  colframe=gray!95,
  width=\linewidth,
  fontupper=\ttfamily\scriptsize,
  before upper={\setlength{\parskip}{0.7\baselineskip}\setlength{\parindent}{0pt}}
]
You will be given:
- A golden fact (a specific claim or statement)
- A response generated by a language model

Your task is to determine whether the golden fact is **explicitly stated or clearly implied** in the response.

Only respond with:
- True — if the fact is stated or clearly implied
- False — if the fact is missing, contradicted, or too vague

\#\#\# Golden Fact:
[Golden Fact]

\#\#\# LLM Response:
[LLM Response]

Output:
\end{tcolorbox}
Parameters: Temperature: 0, Max Tokens: 10

\end{document}